\documentclass[10pt,twocolumn,letterpaper]{article}

\usepackage[pagenumbers]{cvpr}
\usepackage[margin=1in]{geometry}
\usepackage[T1]{fontenc}
\usepackage{amsmath,amssymb,booktabs,xcolor}
\usepackage{cuted}

\definecolor{cvprblue}{rgb}{0.21,0.49,0.74}
\usepackage[breaklinks,colorlinks,allcolors=cvprblue]{hyperref}

\hypersetup{pdftitle={OceanFix: Reference-Guided Diffusion for Sparse-View Underwater 3D Reconstruction}, pdfauthor={Gen Li, Maoxin Chen, Yao Xiao, Xinyang Li, Yu Pan}}

\title{OceanFix: Reference-Guided Diffusion\\for Sparse-View Underwater 3D Reconstruction}

\author{
Gen Li\textsuperscript{1}\quad Maoxin Chen\textsuperscript{2}\quad Yao Xiao\textsuperscript{1}\quad Xinyang Li\textsuperscript{1}\quad Yu Pan\textsuperscript{1}\\[3pt]
{\normalsize \textsuperscript{1}Zhejiang University\qquad \textsuperscript{2}Jiangsu Normal University}}

\begin{document}
\maketitle
\begin{strip}
\centering
\captionsetup{type=figure}
\includegraphics[width=\linewidth]{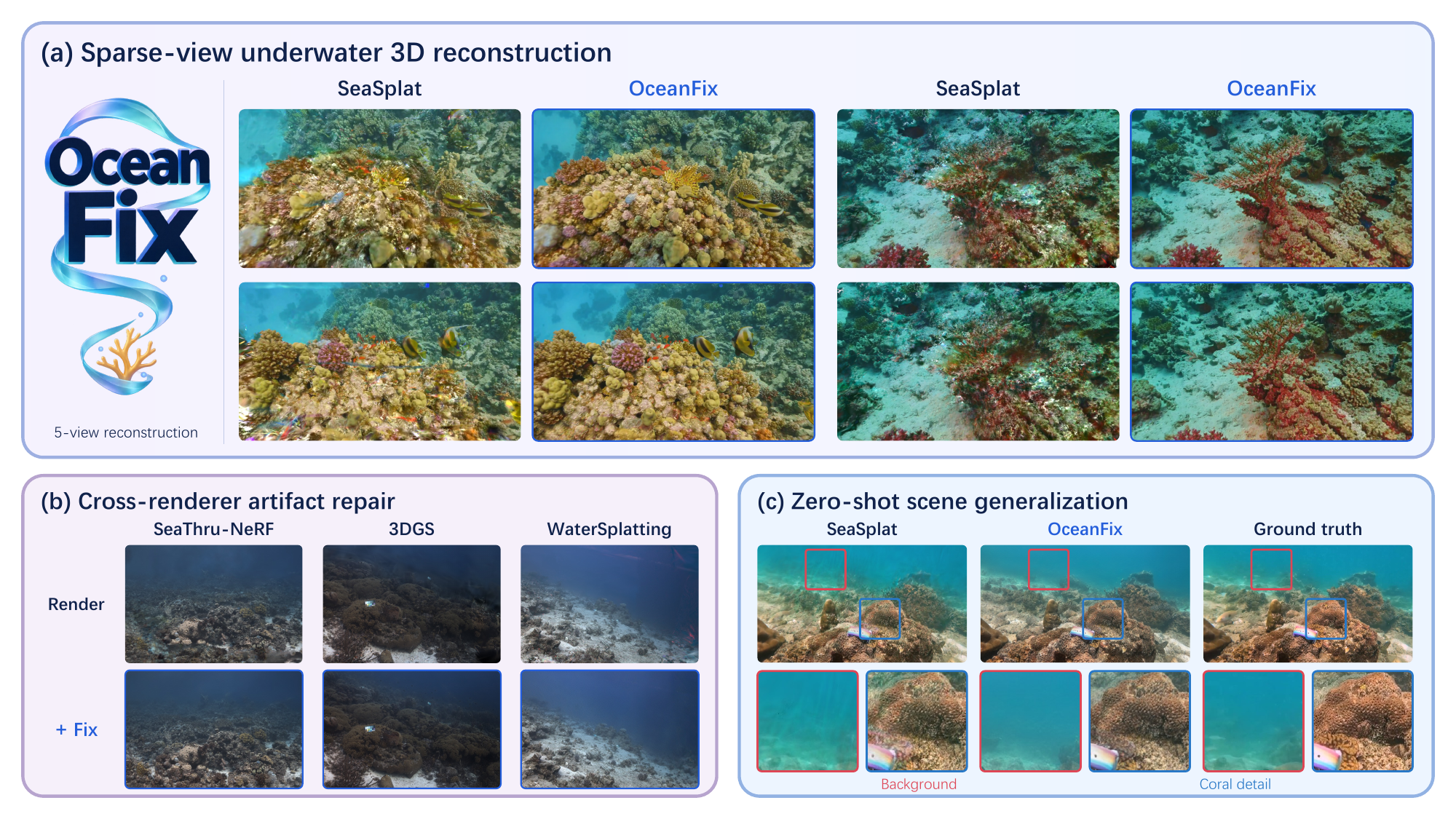}
\caption{\textbf{OceanFix for sparse-view underwater 3D reconstruction.}
\textbf{(a)} From only five observed views, OceanFix reduces rendering artifacts
and improves scene details compared with SeaSplat.
\textbf{(b)} The same trained fixer repairs outputs from different reconstruction
methods without additional fine-tuning.
\textbf{(c)} Zero-shot generalization to unseen underwater scenes, with enlarged
crops highlighting background appearance and coral details.}
\label{fig:teaser}
\end{strip}

\begin{abstract}
Recent advances in Gaussian splatting have enabled efficient underwater reconstruction, but sparse observations, light attenuation, and backscatter still leave scene regions underconstrained and cause novel-view artifacts. We present OceanFix, a reference-guided framework for sparse-view underwater 3D reconstruction that combines diffusion-based view repair with Gaussian scene refinement. Our repair model follows a two-stage training strategy that separates underwater appearance adaptation from geometric refinement. The first stage learns underwater appearance from reference observations and estimated medium features. The second uses camera poses and reference depth to align reference features, with relative-depth supervision guiding structural repair without requiring target-view depth at inference. We further propose medium-aware distillation that transfers repaired views into the 3D scene according to geometric repair demand and channel-specific optical observability. Through iterative rendering, repair, and distillation, OceanFix refines underconstrained regions while retaining supervision from the observed views. Experiments on underwater datasets demonstrate that OceanFix improves rendering quality in sparse-view 3D reconstruction, reducing visible artifacts and recovering finer scene details.
\end{abstract}

\section{Introduction}
\label{sec:introduction}

Reconstructing underwater scenes from a small number of images requires recovering
both scene structure and its appearance through water. Neural radiance fields
(NeRF)~\cite{mildenhall2020nerf,barron2022mipnerf360,mueller2022instant,barron2023zipnerf}
and 3D Gaussian splatting
(3DGS)~\cite{kerbl2023gaussians,yu2024mipsplatting,lu2024scaffoldgs,guedon2024sugar} have advanced image-based reconstruction and
novel-view synthesis, with 3DGS enabling efficient optimization and real-time
rendering. Their underwater application must additionally account for
wavelength-dependent light attenuation and backscatter, which couple the observed
appearance to the intervening medium~\cite{akkaynak2018underwater,levy2023seathrunerf,yang2025seasplat}. When only a
few viewpoints are available, this coupling compounds the geometric ambiguity
caused by limited coverage. Our goal is to reconstruct a renderable underwater
scene from these sparse observations while reducing artifacts at unobserved
viewpoints.

Recent underwater reconstruction methods explicitly incorporate the medium into
scene modeling. SeaThru-NeRF~\cite{levy2023seathru} jointly learns scene and medium
properties through volumetric rendering, while WaterSplatting~\cite{li2024watersplatting}
combines Gaussian geometry with a separate volumetric medium field. Related
restoration approaches, such as RestorGS~\cite{qiao2025restorgs}, use depth-guided
degradation modeling to separate clear and degraded appearance. These advances
provide an essential account of underwater image formation. However, modeling
the medium does not resolve the lack of geometric observations in sparse-view
reconstruction. Sparse-input methods such as RegNeRF~\cite{niemeyer2022regnerf}
address this ambiguity by regularizing geometry and appearance at unobserved
views, motivating additional constraints beyond fitting the available images.

Generative priors~\cite{liu2023zero1to3,sargent2024zeronvs,
zheng2024free3d,gao2024cat3d,li2026spvc} offer another source of supervision for underconstrained scene
regions. ReconFusion~\cite{wu2024reconfusion} uses diffusion-based novel-view
generation to regularize sparse-view reconstruction, while
Difix3D+~\cite{wu2025difix} repairs rendered views with a single-step diffusion
model and distills them back into the 3D representation. More recently,
Ocean4D~\cite{huang2026ocean4d} combines geometric conditioning with medium-aware
video diffusion for underwater generation from monocular videos. Although these
works demonstrate the value of generative supervision, applying such priors
directly to sparse-view underwater 3D reconstruction remains challenging.
Underwater appearance depends on both scene geometry and light propagation
through the medium, making image repair alone insufficient for reliable scene
refinement. First, reference features are not spatially aligned with the target
view, and transferring their appearance without accounting for geometry and
view-dependent medium effects can introduce structural inconsistencies.
Second, repaired pixels provide unequal supervision for the underlying objects:
attenuation suppresses object signals differently across color channels, while
backscatter can dominate the observed appearance. Effective generative
supervision for underwater reconstruction must therefore account for both
reference-to-target alignment and medium-dependent signal reliability,
directing scene refinement toward regions that require geometric repair and
retain observable object information.

To address these challenges, we present \textbf{OceanFix} as shown in Fig.~\ref{fig:teaser}, a reference-guided
framework for sparse-view underwater 3D reconstruction that combines
diffusion-based view repair with medium-aware Gaussian refinement. We train the repair model in two
stages. The first adapts underwater appearance using reference RGB features and
estimated medium descriptors. The second freezes the appearance parameters and
learns geometry adapters that project reference features into the target view
using camera poses and estimated reference depth. A relative-depth objective
further constrains structural changes during training. This separation allows
geometric adaptation while retaining the learned appearance model, and requires
no target-view depth at inference. Both types of reference information are used
in a single repair pass.

We further introduce a medium-aware distillation mechanism to transfer repaired
views into the Gaussian scene. We combine geometric
repair demand with channel-specific optical observability. Depth dispersion
along rendering rays provides a proxy for local geometric ambiguity, while the
underwater renderer estimates the contribution of object appearance to each color
channel. Their combination assigns additional supervision to regions with greater
repair demand and stronger object visibility. Bounded, normalized weights retain
a baseline contribution across the image. We periodically refresh the repaired
pseudo-views and their weights as the reconstruction evolves, while preserving
supervision from the observed images. The resulting rendering--repair--distillation
loop connects reference-guided image corrections to refinement of the explicit
3D representation. Our contributions are summarized as follows:
\begin{itemize}
  \item We introduce OceanFix, a framework for sparse-view underwater 3D
  reconstruction that integrates reference-guided diffusion repair with iterative
  Gaussian scene refinement.
  \item We develop a two-stage repair model that first adapts underwater appearance
  and then introduces explicit geometric constraints through pose-aligned
  reference features and relative-depth supervision, without requiring target-view
  depth at inference.
  \item We propose a medium-aware distillation mechanism that combines geometric
  repair demand with channel-specific optical observability to weight repaired-view
  supervision during reconstruction.
\end{itemize}

\section{Related Work}
\label{sec:related_work}

\subsection{Underwater 3D Reconstruction}
Underwater reconstruction requires modeling both scene surfaces and participating
media. Physically grounded image formation distinguishes direct-signal
attenuation from backscatter~\cite{akkaynak2018underwater}, forming the basis of
SeaThru-NeRF~\cite{levy2023seathru}. Neural Underwater Scene
Representation~\cite{tang2024nusr} further models unstable in-scattering and moving
objects. WaterSplatting~\cite{li2024watersplatting} and
SeaSplat~\cite{yang2025seasplat} integrate medium modeling with Gaussian
representations for efficient rendering, while RestorGS~\cite{qiao2025restorgs}
uses appearance decoupling and depth-guided degradation modeling for restoration.
UDF~\cite{liu2026udf} extends reconstruction to dynamic scenes with time-varying
geometry, medium effects, and uncertainty-aware optimization.

\subsection{Sparse-View 3D Reconstruction}
Sparse-view reconstruction mitigates ambiguity from limited observations through
learned priors and per-scene regularization. pixelNeRF~\cite{yu2021pixelnerf}
learns image-conditioned radiance fields across scenes, while
RegNeRF~\cite{niemeyer2022regnerf} regularizes geometry and appearance at unobserved
views. FreeNeRF~\cite{yang2023freenerf} controls input frequencies and penalizes
near-camera density. For Gaussian representations, FSGS~\cite{zhu2024fsgs}
combines proximity-guided Gaussian unpooling with geometric regularization,
whereas DNGaussian~\cite{li2024dngaussian} uses monocular depth supervision and
global--local depth normalization. Our underwater setting additionally couples
incomplete geometry with medium-dependent appearance. We constrain diffusion
repair using reference depth and camera poses, then distill repaired views into
the underwater Gaussian scene, complementing direct geometric regularization.

\subsection{3D Reconstruction with Diffusion Priors}
Diffusion priors aid reconstruction through regularization, view generation,
and rendering repair. DiffusioNeRF~\cite{wynn2023diffusionerf} learns an RGB--depth
prior, while ReconFusion~\cite{wu2024reconfusion} and CAT3D~\cite{gao2024cat3d}
generate novel-view supervision. GaussianObject~\cite{yang2024gaussianobject}
and Difix3D+~\cite{wu2025difix} refine 3D representations using repaired
renderings. RI3D~\cite{paliwal2025ri3d} and GSFix3D~\cite{wei2025gsfix3d}
address both artifacts and missing regions, while GSFixer~\cite{yin2025gsfixer}
uses reference-guided video diffusion for consistent repair.
Ocean4D~\cite{huang2026ocean4d} combines geometric conditioning with medium-aware
video diffusion for underwater generation.
OceanFix adapts repair-and-distill to sparse-view underwater reconstruction
through separate appearance and geometry training, and weights repaired-view
supervision by geometric repair demand and channel-specific optical
observability.

\section{Method}
\label{sec:method}
\begin{figure*}[tp]
\centering
\includegraphics[width=\linewidth]{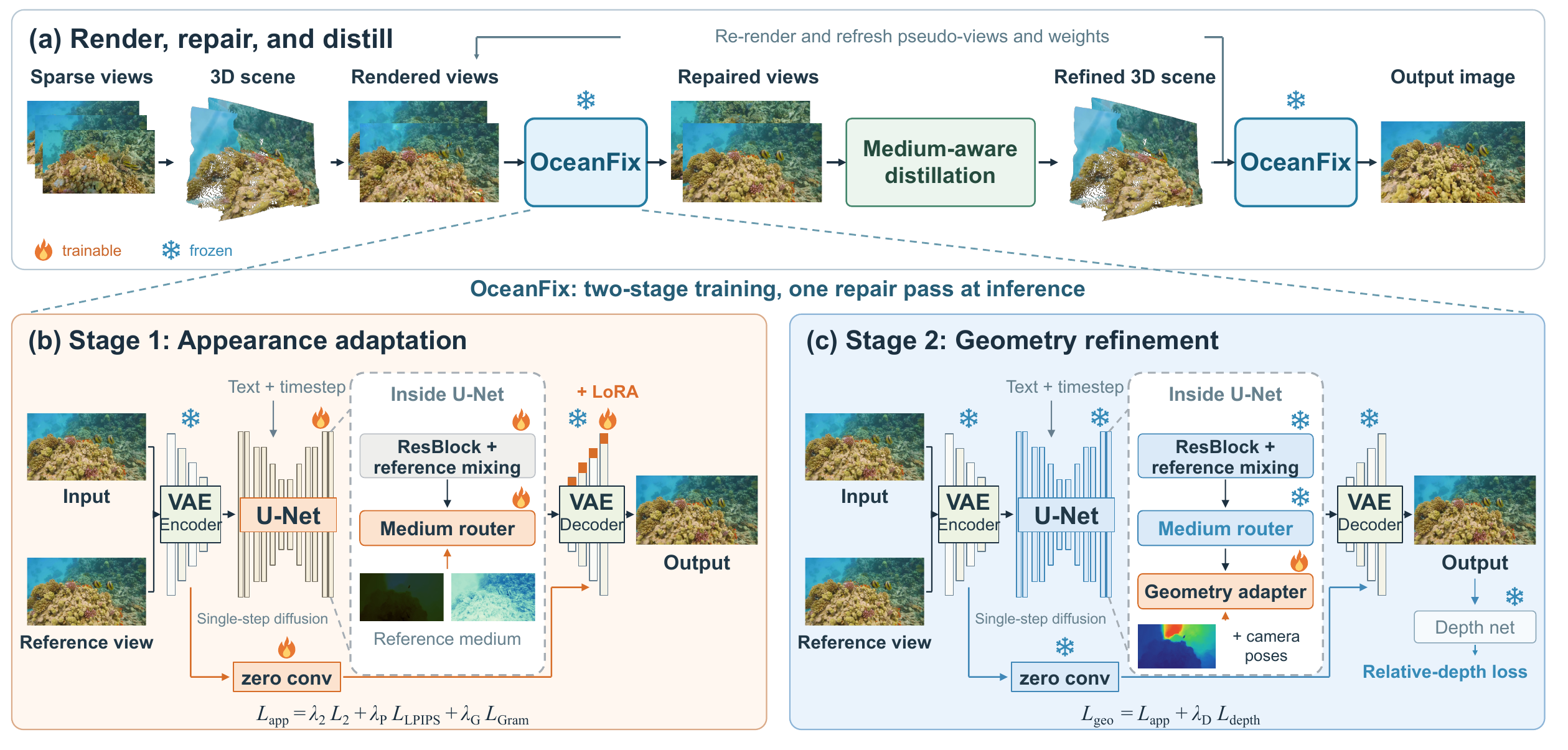}
\caption{\textbf{Overview of OceanFix.}
(a) Rendered views are repaired for medium-aware distillation, with
pseudo-views and weights refreshed during refinement. A final repair
pass produces the output images.
(b) Appearance adaptation learns reference RGB and medium conditioning.
(c) Geometry refinement freezes appearance parameters and trains
pose-aligned geometry adapters with relative-depth supervision.}
\label{fig:method_overview}
\end{figure*}

\subsection{Problem formulation and overview}
Given posed underwater observations $\mathcal{O}=\{(I_i,K_i,P_i)\}$,
where $K_i$ denotes camera intrinsics and $P_i$ a camera-to-world transform,
we reconstruct an underwater scene with Gaussian parameters $\theta$ and
medium parameters $\phi$. For a target camera $\pi_t=(K_t,P_t)$, the renderer
produces $\widetilde I_t=\mathcal{R}(\theta,\phi;\pi_t)$.
The fixer uses this rendering, a reference observation $I_r$, reference
auxiliary maps, and both cameras to predict a repaired underwater image
$\widehat I_t$, preserving the underwater appearance.

The reference auxiliary maps comprise axial depth $D_r$, accumulated object
opacity $A_r$, medium contribution $B_r$, and medium fraction $f_r$, all
rendered from the initial reconstruction. Here, $f_r$ measures the medium's
share of the total RGB signal. Target RGB ground truth and target depth
are not required at inference.

The fixer is trained in two stages: underwater appearance adaptation followed
by geometry-adapter training. At inference, it repairs each view in a single
pass and remains frozen throughout the iterative rendering--repair--distillation
process used to refine the scene.

\subsection{Reference-conditioned medium appearance}
\label{sec:appearance}
We adapt underwater appearance by adding a reference router to the pretrained
single-step Difix model~\cite{wu2025difix}, retaining its paired target/reference
backbone. At four U-Net down-block resolutions, target features provide queries,
while reference RGB features and estimated medium descriptors $b_r=[B_r,f_r]$
form the keys. Values encode the medium descriptors, allowing spatially varying
appearance retrieval without requiring pixelwise correspondence between views.
The attention weights are
\begin{equation}
 p_{ij} = \operatorname{softmax}_j
 \left(\frac{q_i^\top k_j}{\sqrt{d_h}}+\log\rho_j\right),
 \label{eq:water_attention}
\end{equation}
where $q_i$ and $k_j$ are the target query and reference key, and $d_h$ is the
per-head feature dimension. The support $\rho_j=f_{r,j}v_{r,j}^{\mathrm{image}}$
combines estimated medium fraction with image validity, masking unsupported
tokens. Retrieved descriptors are scaled by the maximum reference support and
injected through a learned sigmoid gate with a zero-initialized residual
projection.

During this stage, we update the U-Net, reference router, and VAE LoRA and skip
parameters, with geometry adapters disabled. The appearance objective combines
pixel reconstruction, perceptual similarity, and feature-correlation matching:
\begin{equation}
 \mathcal{L}_{\mathrm{RGB}}=
 \mathcal{L}_2+
 \lambda_P\mathcal{L}_{\mathrm{LPIPS}}+
 \lambda_G\mathcal{L}_{\mathrm{Gram}}.
 \label{eq:rgb_loss}
\end{equation}
Each term averages losses over the target and reference outputs, supervised by
their corresponding ground-truth images. Here $\mathcal{L}_2$ is image MSE,
$\mathcal{L}_{\mathrm{LPIPS}}$ measures perceptual similarity~\cite{zhang2018lpips},
and $\mathcal{L}_{\mathrm{Gram}}$ compares VGG feature correlations between each
prediction and its paired target. We set $\lambda_P=1$ and switch $\lambda_G$
from zero to one after global step 2000. Image losses exclude letterbox padding.

\subsection{Pose-aligned reference geometry}
\label{sec:geometry}
In the second stage, we freeze all appearance-stage parameters and train four
geometry adapters to align reference features with the target view. Reference
depth $D_r$ is rendered from SeaSplat, sharing the coordinate scale of the
camera poses. Given the relative pose $(R_{tr},t_{tr})$ from $P_t^{-1}P_r$, we
project a reference pixel into the target view:
\begin{equation}
 \begin{aligned}
 X_r &= D_r(u_r)K_r^{-1}\widetilde u_r,\\
 u_t &= \pi\!\left(K_t(R_{tr}X_r+t_{tr})\right),
 \end{aligned}
 \label{eq:warp}
\end{equation}
where $\pi$ denotes perspective division. Forward z-buffering retains the
nearest valid reference projection at each target pixel. Its support weight
$m_t=A_r(1-f_r)$ combines reference object opacity and medium fraction,
transported to the target view; unsupported pixels receive zero weight.

Each adapter fuses the aligned reference features, target features, a relative
camera embedding, projected depth, and the support map. The camera embedding
encodes relative pose and intrinsics, with translation normalized by reference
depth. A learned gate injects the resulting feature residual, weighted by $m_t$,
into the target branch. Zero-initialized residual projections preserve the
appearance-stage output when the adapters are first enabled.

\noindent\textbf{Geometry supervision. }We regularize the repaired structure using a frozen Depth Anything V2 Large
estimator $E$~\cite{yang2024depth}. Let $\widehat d$ and $d^\star$ denote
$E(\widehat I_t)$ and $E(I_t^\star)$, each standardized using its support-weighted
mean and standard deviation. These are relative inverse-depth estimates from
the repaired image and its training target. The depth loss is
\begin{equation}
 \mathcal{L}_{\mathrm{depth}}=
 \left\langle\ell_{\mathrm{SL1}}(\widehat d,d^\star)\right\rangle_m,
 \label{eq:depth_loss}
\end{equation}
where $\langle\cdot\rangle_m$ denotes the support-weighted pixel average and
$\ell_{\mathrm{SL1}}$ is Smooth L1. The second-stage objective is
\begin{equation}
 \mathcal{L}_{\mathrm{stage2}}=
 \mathcal{L}_{\mathrm{RGB}}+
 \lambda_D\mathcal{L}_{\mathrm{depth}},
 \label{eq:stage2_loss}
\end{equation}
with $\lambda_D=0.05$. Gradients pass through $E$ to the repaired image while
its parameters remain frozen. This objective constrains relative structure
within projected reference support. Neither target-view depth nor the depth
estimator is required at inference.

\subsection{Demand-weighted underwater distillation}
\label{sec:distillation}
Following the render--repair--update framework~\cite{wu2025difix},
we refine a SeaSplat scene~\cite{yang2025seasplat} using repaired
pseudo-views, weighted by geometric repair demand and optical observability.

\noindent\textbf{Geometric repair demand.} Let $a_i(p)$ denote the alpha-compositing contribution of Gaussian $i$,
excluding water attenuation~\cite{kerbl2023gaussians}, and
$A(p)=\sum_i a_i(p)$ its accumulated opacity.
Using normalized contributions $a_i/A$, we compute the mean $\mu_z$
and standard deviation $\sigma_z$ of camera-space Gaussian depths.
Their relative dispersion $s(p)=\sigma_z(p)/\mu_z(p)$ defines
\begin{equation}
 G(p)=\frac{s(p)}{s(p)+s_0},
 \label{eq:demand}
\end{equation}
where $s_0$ is the median valid dispersion with a positive floor.
Invalid pixels receive $G=0$. This score indicates relative repair
priority rather than calibrated artifact probability.

\noindent\textbf{Optical observability.} SeaSplat applies channel-specific transmission $T_c$ after color compositing.
We define
\begin{equation}
 K_c(p)=A(p)T_c(p),
 \label{eq:observability}
\end{equation}
which measures object-color sensitivity at fixed geometry and medium,
excluding surface brightness and additive backscatter.

\noindent\textbf{Weighted supervision and refresh.} We combine these signals into
\begin{equation}
 w_{p,c}=\frac{1+G(p)K_c(p)^2}{Z},
 \label{eq:weights}
\end{equation}
where $Z$ normalizes the mean weight to one.
The weights retain a supervision floor while emphasizing depth-dispersed
regions with observable object contributions.

At round $r$, the frozen fixer produces repaired targets $Y_r$.
Scene optimization minimizes
\begin{equation}
 \mathcal{L}_{\mathrm{scene}}=
 \mathcal{L}_{\mathrm{obs}}+
 \lambda_1\mathcal{L}_{1,w}+
 \lambda_S\mathcal{L}_{\mathrm{SSIM}},
 \label{eq:distill_loss}
\end{equation}
where $\mathcal{L}_{\mathrm{obs}}$ follows the SeaSplat observed-view objective.
The pseudo-view terms compare renderings with $Y_r$ using weighted L1
and unweighted one-minus-SSIM losses.

\subsection{Training Data Curation}
\label{sec:data_curation}

\begin{figure}[tbp]
\centering
\includegraphics[width=\linewidth]{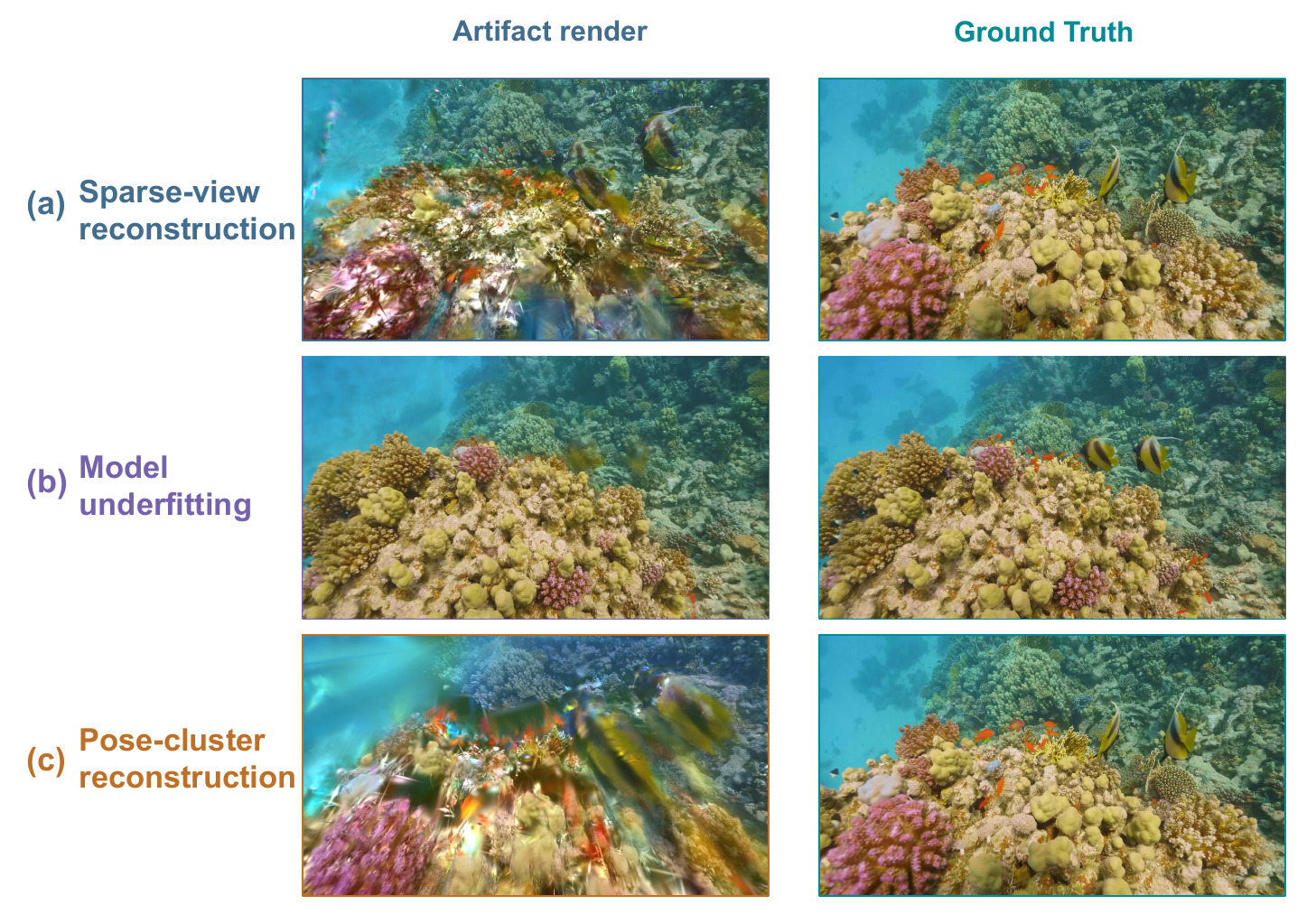}
\caption{\textbf{Training-pair construction.} Sparse-view reconstruction, underfit checkpoints, and complementary pose clusters generate degraded renders paired with allowed GroundTruth.}
\label{fig:data_curation}
\end{figure}

We use SeaSplat~\cite{yang2025seasplat} to construct degraded-rendering,
observed-target, and reference-image triplets on NUSR and SeaThru-NeRF.
We generate degradations through three strategies:
\textit{sparse-view reconstruction} using three or five uniformly sampled
views for 30,000 iterations;
\textit{underfitting} using checkpoints at 12,000, 16,000, and 20,000 iterations
to render approximately 20\% held-out views; and
\textit{pose-cluster reconstruction}, which partitions cameras into two
clusters and trains on each cluster to render the other.
Each target observation is excluded from fitting its corresponding
reconstruction. We select the nearest permitted reference using camera
translation and rotation distances, and obtain its depth, opacity, and
medium descriptors from the same SeaSplat checkpoint.
Camera parameters and scene-scale reference depth support geometric
projection, while target relative depth provides training-only supervision.
We use four leave-one-scene-out folds per dataset, constructing ordinary
training pairs from the other three scenes.
Evaluation targets never serve as supervision or references, and
BVI-Coral remains excluded from training.
Figure~\ref{fig:data_curation} illustrates the resulting pairs.

\section{Experiments}
\label{sec:experiments}

\subsection{Experimental setup}
\label{sec:experimental_setup}
\paragraph{Datasets.}
We conduct experiments on three underwater datasets:
NUSR~\cite{tang2024nusr},
SeaThru-NeRF~\cite{levy2023seathru}, and
BVI-Coral~\cite{anantrasirichai2024bvicoral,wang2025uwgs}.
NUSR captures underwater environments with scattering effects and moving
objects. SeaThru-NeRF provides real underwater imagery with varying attenuation
and backscatter, while BVI-Coral captures coral reef environments.
We use 3 and 5 input images for reconstruction on NUSR and SeaThru-NeRF,
and 5\% of the registered frames on BVI-Coral.
\begin{figure*}[t]
\centering
\includegraphics[width=\linewidth]{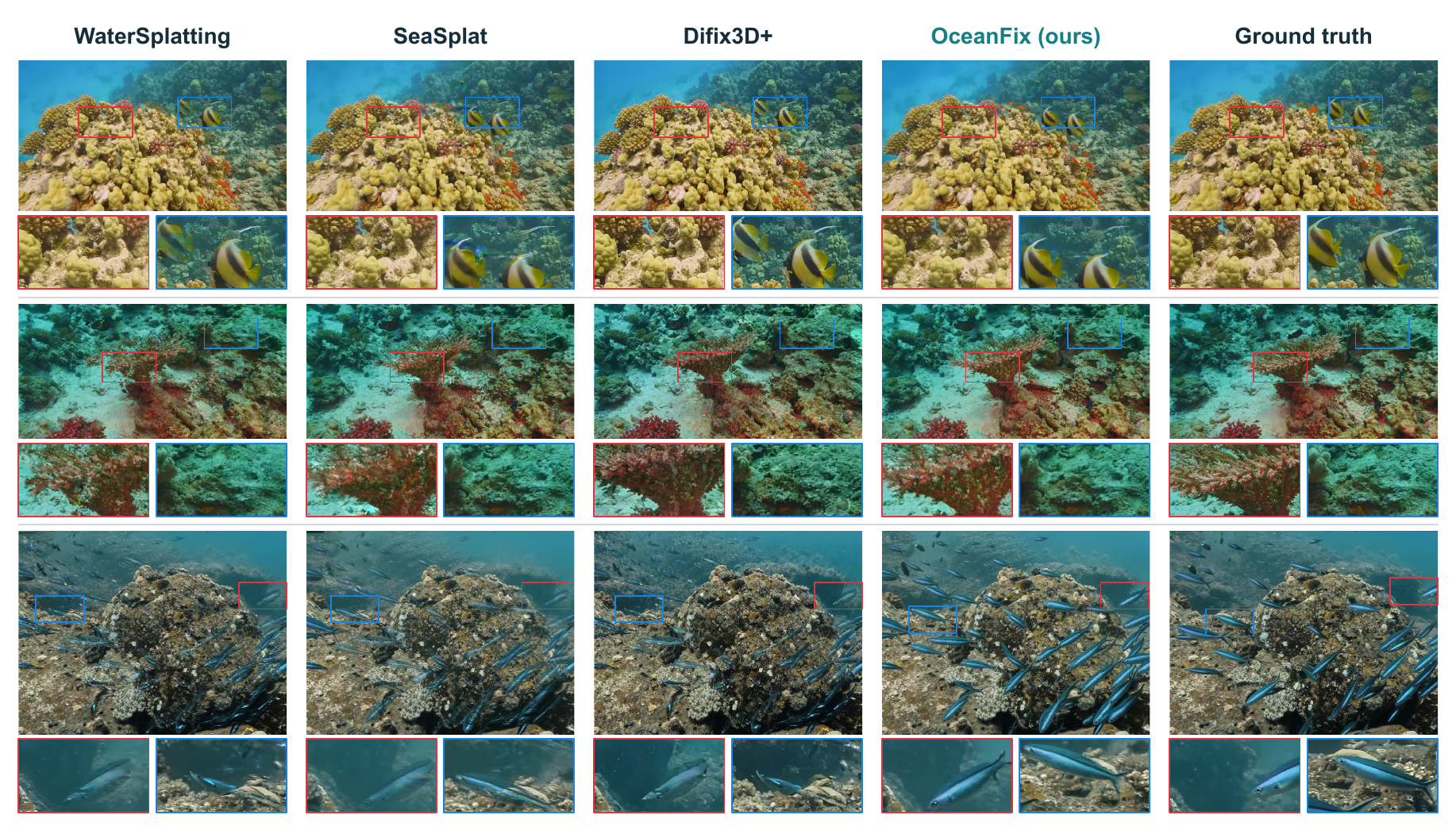}
\caption{\textbf{Qualitative comparison on NUSR.} OceanFix achieves the best visual quality, with clearer foreground details and more consistent underwater backgrounds. Difix3D+ is fine-tuned on our underwater training pairs.}
\label{fig:qualitative_nusr}
\end{figure*}

\begin{figure*}[t]
\centering
\includegraphics[width=\linewidth]{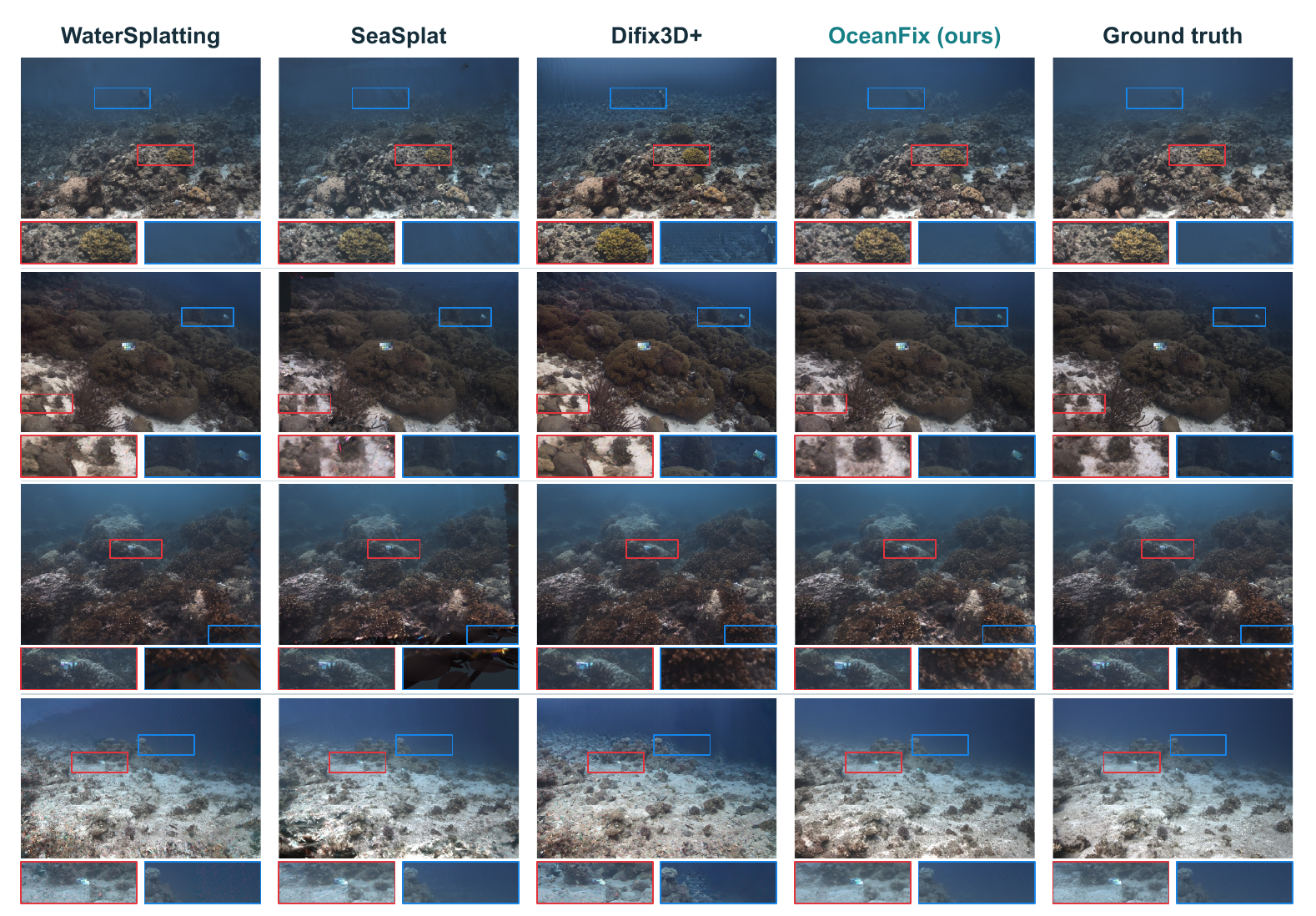}
\caption{\textbf{Qualitative comparison on SeaThru-NeRF.} OceanFix produces the most visually faithful reconstructions, preserving finer scene structures and more natural underwater appearance. Difix3D+ is fine-tuned on our underwater training pairs.}
\label{fig:qualitative_seathrunerf}
\end{figure*}

\begin{figure}[tbp]
\centering
\includegraphics[width=\linewidth]{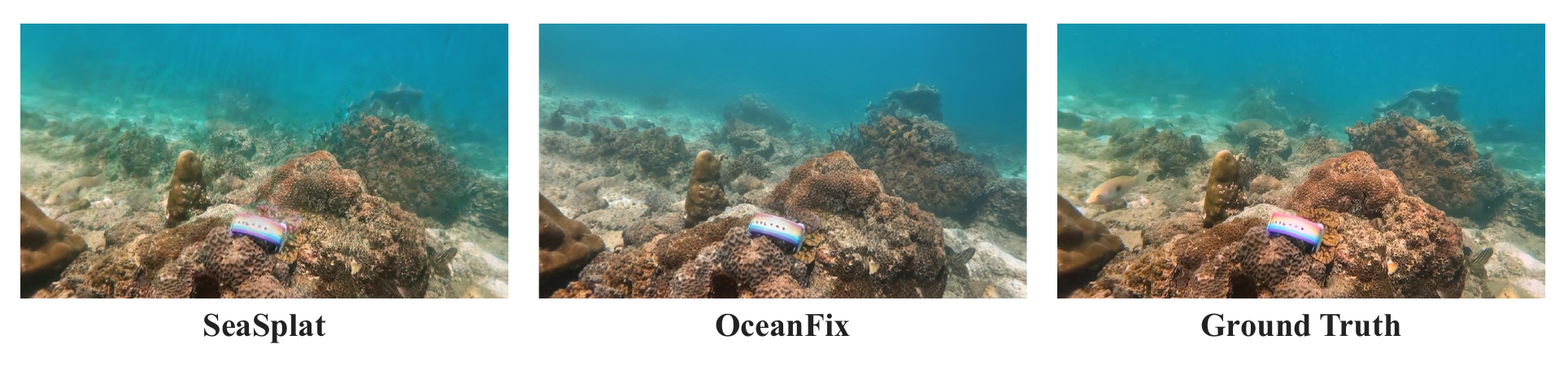}
\caption{Qualitative results on BVI-Coral using 5\% of the images as input for sparse-view reconstruction.}
\label{fig:qualitative_bvi}
\end{figure}

\begin{figure}[tbp]
\centering
\includegraphics[width=\linewidth]{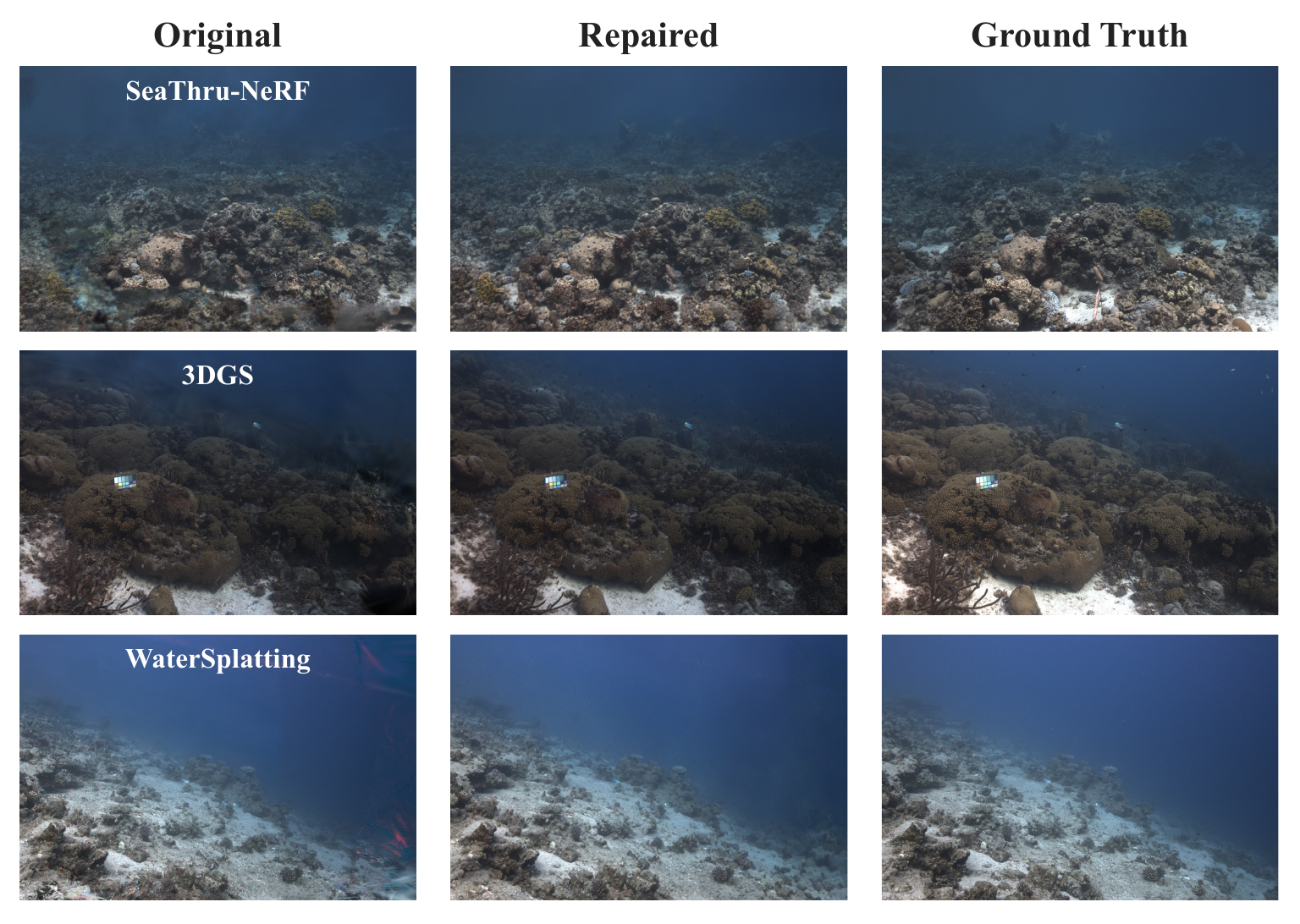}
\caption{\textbf{Qualitative results across different reconstruction
renderers on SeaThru-NeRF.} The same frozen fixer, trained on
SeaSplat pairs, is applied to all methods with identical
reference conditions and no additional fine-tuning.}
\label{fig:cross_renderer}
\end{figure}

\begin{table}[tbp]
\centering
\caption{Effect of repair and distillation on the IUI3-RedSea scene
of SeaThru-NeRF under 5-view reconstruction.
Distillation uses repaired pseudo-views to refine SeaSplat;
the final Fix is applied to the refined scene renderings.}
\label{tab:ablation_fixer}
\small
\setlength{\tabcolsep}{3pt}
\begin{tabular*}{\columnwidth}{@{\extracolsep{\fill}}lccc@{}}
\toprule
Variant & PSNR$\uparrow$ & SSIM$\uparrow$ & FID$\downarrow$ \\
\midrule
Input & 21.69 & 0.7344 & 41.64 \\
Direct Fix &  \underline{23.57} &  \underline{0.7575} & \underline{17.16} \\
Distillation & 21.93 & 0.7491 & 40.98 \\
\textbf{Distillation + Fix} & \textbf{23.64} & \textbf{0.7579} & \textbf{16.40} \\
\bottomrule
\end{tabular*}
\end{table}

\paragraph{Baselines.}
We compare OceanFix with three groups of baselines.
(1) NeRF-based methods include Zip-NeRF~\cite{barron2023zipnerf}
and SeaThru-NeRF~\cite{levy2023seathru}.
(2) 3DGS-based underwater reconstruction methods include
SeaSplat~\cite{yang2025seasplat} and
WaterSplatting~\cite{li2024watersplatting}.
(3) Diffusion-assisted refinement methods include
Difix3D+~\cite{wu2025difix} and Difix3D+ (fine-tuned).
The latter uses a Difix repair model fine-tuned on our underwater training
pairs, enabling comparison with domain-adapted diffusion repair.

\noindent\textbf{Metrics.}
We evaluate rendering quality using PSNR, SSIM~\cite{wang2004ssim},
LPIPS~\cite{zhang2018lpips}, and FID~\cite{heusel2017fid}.
PSNR and SSIM measure pixel fidelity and structural similarity, respectively,
while LPIPS measures perceptual distance and FID assesses distributional
similarity. Higher PSNR and SSIM and lower LPIPS and FID indicate better quality.

\subsection{Sparse-view reconstruction evaluation}
\label{sec:main_evaluation}
Tab.~\ref{tab:main_nusr} and~\ref{tab:main_seathrunerf} show that
OceanFix achieves the best overall reconstruction quality under both
sparse-view settings. Despite underwater fine-tuning, adapted Difix
remains less effective in handling complex underwater appearance,
resulting in lower reconstruction quality. The qualitative comparisons
in Fig.~\ref{fig:qualitative_nusr}
and~\ref{fig:qualitative_seathrunerf} further support these results:
OceanFix produces visually cleaner renderings with better-preserved
scene details and more consistent underwater appearance.

\subsection{Robustness Across Reconstruction Renders}
\label{sec:cross_renderer}
\begin{table}[tbp]
\centering
\caption{Robustness to different reconstruction renderings on
SeaThru-NeRF. The same frozen fixer, trained on Seasplat pairs,
is applied to all methods with identical reference conditions and
no additional fine-tuning.}
\label{tab:cross_renderer}
\small
\setlength{\tabcolsep}{2pt}
\begin{tabular*}{\columnwidth}{@{\extracolsep{\fill}}llccc@{}}
\toprule
Renderer & Output & PSNR$\uparrow$ & SSIM$\uparrow$ & LPIPS$\downarrow$ \\
\midrule
SeaThru-NeRF & Original & 19.26 & 0.5575 & 0.3755 \\
             & Repaired & \textbf{19.34} & \textbf{0.5797} & \textbf{0.3513} \\
\addlinespace[3pt]
3DGS & Original & 21.30 & 0.7299 & 0.3376 \\
     & Repaired & \textbf{21.99} & \textbf{0.7413} & \textbf{0.2833 }\\
\addlinespace[3pt]
WaterSplatting & Original & 22.70 & 0.7306 & 0.2686 \\
              & Repaired & \textbf{22.79} & \textbf{0.7336} & \textbf{0.2412} \\
\bottomrule
\end{tabular*}
\end{table}

We apply the frozen OceanFix fixer trained on SeaSplat pairs
to SeaThru-NeRF, 3DGS, and WaterSplatting renderings under 5-view
reconstruction on SeaThru-NeRF dataset. All methods use identical
reference conditions without additional fine-tuning.
Tab.~\ref{tab:cross_renderer} shows improved PSNR, SSIM, and LPIPS
across all three renderers, supporting transfer beyond the training
renderer. Fig.~\ref{fig:cross_renderer} provides the corresponding
qualitative comparisons.
\begin{table*}[t]
\centering
\caption{Main comparison on NUSR under 3-view and 5-view settings,
averaged equally across four scenes. $\dagger$ denotes underwater
fine-tuning. Best and second-best results are shown in bold and
underlined, respectively.}
\label{tab:main_nusr}
\small
\setlength{\tabcolsep}{3.5pt}
\begin{tabular}{@{}lcccccccc@{}}
\toprule
& \multicolumn{4}{c}{3-view reconstruction}
& \multicolumn{4}{c}{5-view reconstruction} \\
\cmidrule(lr){2-5}\cmidrule(l){6-9}
Method
& PSNR$\uparrow$ & SSIM$\uparrow$ & LPIPS$\downarrow$ & FID$\downarrow$
& PSNR$\uparrow$ & SSIM$\uparrow$ & LPIPS$\downarrow$ & FID$\downarrow$ \\
\midrule
Zip-NeRF
& 16.58 & 0.4639 & 0.5209 & 86.93
& 18.49 & 0.5643 & 0.3813 & 71.03 \\
SeaThru-NeRF
& 13.46 & 0.3302 & 0.6693 & 103.5
& 13.78 & 0.3510 & 0.6091 & 73.52 \\
3D Gaussian Splatting
& 19.40 & 0.5748 & 0.3501 & 55.65
& 22.82 & 0.7178 & 0.2354 & 46.15 \\
SeaSplat
& \underline{20.27} & \underline{0.6097} & \underline{0.3113} & 44.89
& \underline{23.14} & \underline{0.7274} & \underline{0.2296} & \underline{39.11} \\
WaterSplatting
& 17.48 & 0.5257 & 0.3958 & 77.20
& 19.31 & 0.6258 & 0.3446 & 150.5 \\
Difix3D+
& 17.13 & 0.4649 & 0.4356 & 64.75
& 18.25 & 0.5276 & 0.3866 & 104.6 \\
Difix3D+$^{\dagger}$
& 19.01 & 0.5469 & 0.3648 & \underline{33.38}
& 19.94 & 0.6298 & 0.2995 & 42.75 \\
\textbf{OceanFix (ours)}
& \textbf{20.67} & \textbf{0.6198} & \textbf{0.3090} & \textbf{30.79}
& \textbf{23.23} & \textbf{0.7326} & \textbf{0.2283} & \textbf{25.81} \\
\bottomrule
\end{tabular}
\end{table*}

\begin{table*}[t]
\centering
\caption{Main comparison on SeaThru-NeRF under 3-view and 5-view settings,
averaged equally across four scenes. $\dagger$ denotes underwater
fine-tuning. Best and second-best results are shown in bold and
underlined, respectively.}
\label{tab:main_seathrunerf}
\small
\setlength{\tabcolsep}{3.5pt}
\begin{tabular}{@{}lcccccccc@{}}
\toprule
& \multicolumn{4}{c}{3-view reconstruction}
& \multicolumn{4}{c}{5-view reconstruction} \\
\cmidrule(lr){2-5}\cmidrule(l){6-9}
Method
& PSNR$\uparrow$ & SSIM$\uparrow$ & LPIPS$\downarrow$ & FID$\downarrow$
& PSNR$\uparrow$ & SSIM$\uparrow$ & LPIPS$\downarrow$ & FID$\downarrow$ \\
\midrule
Zip-NeRF
& 18.51 & 0.5640 & 0.4250 & 64.45
& 21.37 & 0.6505 & 0.2767 & 33.68 \\
SeaThru-NeRF
& 18.01 & 0.4884 & 0.4833 & 59.37
& 19.26 & 0.5575 & 0.3755 & 52.02 \\
3D Gaussian Splatting
& 17.45 & 0.5809 & 0.4630 & 158.1
& 21.30 & 0.7299 & 0.3376 & 76.03 \\
SeaSplat
& 18.42 & \underline{0.6242} & 0.3675 & 89.95
& 22.38 & \underline{0.7621} & \underline{0.2344} & 36.69 \\
WaterSplatting
& 19.68 & 0.5993 & 0.3727 & 59.76
& 22.70 & 0.7306 & 0.2686 & 30.42 \\
Difix3D+
& 18.95 & 0.5320 & 0.4169 & 47.36
& 21.38 & 0.6245 & 0.3414 & 35.33 \\
Difix3D+$^{\dagger}$
& \underline{19.82} & 0.6125 & \underline{0.3190} & \underline{30.50}
& \underline{22.90} & 0.7316 & 0.2409 & \underline{23.12} \\
\textbf{OceanFix (ours)}
& \textbf{19.86} & \textbf{0.6387} & \textbf{0.3187} & \textbf{28.14}
& \textbf{23.21} & \textbf{0.7659} & \textbf{0.2274} & \textbf{22.96} \\
\bottomrule
\end{tabular}
\end{table*}

\begin{table}[tbp]
\centering
\caption{Quantitative results on BVI-Coral using 5\% of the images as input for sparse-view reconstruction.}
\label{tab:main_bvi}
\small
\setlength{\tabcolsep}{2.5pt}
\begin{tabular*}{\columnwidth}{@{\extracolsep{\fill}}lcccc@{}}
\toprule
Method & PSNR$\uparrow$ & SSIM$\uparrow$ & LPIPS$\downarrow$ & FID$\downarrow$ \\
\midrule
Seasplat & 18.43 & 0.5952 & 0.3115 & 23.25 \\
OceanFix & \textbf{18.56} & \textbf{0.5963} & \textbf{0.3059} & \textbf{18.80} \\
\bottomrule
\end{tabular*}
\end{table}

\subsection{Zero-Shot Generalization to BVI-Coral}
\label{sec:zero_shot}

We evaluate OceanFix on BVI-Coral at 5\% input coverage without
fine-tuning the fixer on this dataset. As shown in
Tab.~\ref{tab:main_bvi}, OceanFix improves all four metrics over
SeaSplat, with modest PSNR and SSIM gains and a reduction in FID
from 23.25 to 18.80. Fig.~\ref{fig:qualitative_bvi} provides the corresponding visual comparison. The fixer remains frozen, using observed reference images and auxiliary
maps from the initial reconstruction to condition its predictions.
These results support the zero-shot transfer of the fixer to BVI-Coral.

\subsection{Ablation Studies}
\label{sec:ablations}

\begin{table}[tbp]
\centering
\caption{Component ablations on the Panama scene of SeaThru-NeRF
under 5-view reconstruction. }
\label{tab:ablation_components}
\small
\setlength{\tabcolsep}{3pt}
\begin{tabular*}{\columnwidth}{@{\extracolsep{\fill}}lccc@{}}
\toprule
Variant & PSNR$\uparrow$ & SSIM$\uparrow$ & LPIPS$\downarrow$ \\
\midrule
w/o RGB reference   & 21.98 & 0.7259 & 0.2728 \\
w/o medium features & 22.07 & 0.7261 & 0.2613 \\
w/o geometry        & 22.69 & 0.7416 & 0.2597 \\
\midrule
Uniform weights       & 22.741 & 0.7438 & 0.2557 \\
Geometry-only weights & 22.746 & 0.7439 & 0.2556 \\
\textbf{OceanFix}      & \textbf{22.756} & \textbf{0.7439} & \textbf{0.2551} \\
\bottomrule
\end{tabular*}
\end{table}

\paragraph{Effect of repair and distillation.}
Tab.~\ref{tab:ablation_fixer} evaluates the contributions of repair
and distillation. Direct repair substantially improves image quality,
while distillation improves scene renderings without a final Fix pass.
Combining both achieves the best results across all three metrics,
demonstrating their complementary benefits. The improvement before
final repair indicates that distillation transfers useful supervision
from repaired images into the underlying scene representation.

\paragraph{Effect of individual components.}
Tab.~\ref{tab:ablation_components} reports component ablations on
the Panama scene of SeaThru-NeRF under 5-view reconstruction.
Removing RGB references, medium features, or geometric guidance
reduces repair quality, with the largest degradation occurring
without RGB references.
For distillation, the full weighting scheme provides small improvements
in PSNR and LPIPS over uniform and geometry-only weights.
OceanFix achieves the highest PSNR and lowest LPIPS, while matching
the best SSIM.

\section{Conclusion}
\label{sec:conclusion}
We presented OceanFix, a reference-guided framework for sparse-view underwater 3D reconstruction. It combines two-stage diffusion repair, guided by underwater appearance and reference geometry, with medium-aware distillation into Gaussian scenes. Pose-aligned reference features and relative-depth supervision provide geometric guidance without requiring target-view depth at inference. During distillation, geometric repair demand and optical observability weight repaired-view supervision, while observed-view supervision and periodic pseudo-view refresh support scene refinement. Experiments on NUSR and SeaThru-NeRF demonstrate improved scene rendering under sparse observations.

\section*{LLM Usage Statement}
We used OpenAI ChatGPT and Codex to assist with drafting, translating, and editing manuscript text, including related-work prose and reference lookup; developing and running training, evaluation, and efficiency-measurement code; analyzing experimental results; and preparing figures and visualizations under the authors' direction.
All reported results, scientific claims, and conclusions were reviewed and verified by the authors, who take full responsibility for the submission.

\end{document}